\documentclass[runningheads]{llncs}
\usepackage[T1]{fontenc}
\usepackage{graphicx}
\usepackage[misc]{ifsym}
\usepackage{multirow}
\usepackage{paralist}
\usepackage{color}
\usepackage{amsmath}
\usepackage{amsfonts}
\usepackage{booktabs}
\usepackage{subcaption}
\usepackage{caption}
\usepackage{float}
\usepackage{algorithmic}
\usepackage[ruled]{algorithm2e}
\newcommand{\corr}{(\Letter)}
\newcommand{\name}{HetCAN}
\usepackage{mwe}

\begin{document}

\title{HetCAN: A Heterogeneous Graph Cascade Attention Network with Dual-Level Awareness}

\titlerunning{HetCAN: Heterogeneous Graph Cascade Attention Network}

\author{Zeyuan Zhao\inst{1} \and
Qingqing Ge\inst{1} \and
Anfeng Cheng \inst{2} \and
Yiding Liu \inst{2} \and \\
Xiang Li\inst{1} \corr \and
Shuaiqiang Wang \inst{2} 
}

\authorrunning{Z. Zhao et al.}

\institute{
School of Data Science and Engineering, East China Normal University, China\\
\email{\{zeyuanzhao, qingqingge\}@stu.ecnu.edu.cn} \\
\email{\{xiangli\}@dase.ecnu.edu.cn}
\and
Baidu Inc., China\\
\email{\{anfcheng2, liuyiding.tanh, shqiang.wang\}@gmail.com}
}

\maketitle              

\begin{abstract}
Heterogeneous graph neural networks (HGNNs) have recently shown impressive capability in modeling heterogeneous graphs that are ubiquitous in real-world applications. 
Most existing methods for heterogeneous graphs mainly learn node embeddings by stacking multiple convolutional or attentional layers, which can be considered as capturing the high-order information from node-level aspect. 
However, different types of nodes in heterogeneous graphs have diverse features,
it is also necessary to capture interactions among node features, namely the high-order information from feature-level aspect.
In addition, 
most methods first align node features by mapping them into one same low-dimensional space, while they may lose some type information of nodes in this way.
To address these problems, 
in this paper,
we propose a novel \textbf{Het}erogeneous graph \textbf{C}ascade \textbf{A}ttention \textbf{N}etwork ({\name}) composed of multiple cascade blocks.
Each cascade block includes two components, the type-aware encoder and the dimension-aware encoder.
Specifically, the type-aware encoder compensates for the loss of node type information and aims to make full use of graph heterogeneity.
The dimension-aware encoder 
is able to learn the feature-level high-order information by capturing the interactions among node features.
With the assistance of these components, {\name} can comprehensively encode information of node features, graph heterogeneity and graph structure in node embeddings.
Extensive experiments demonstrate the superiority of {\name} over advanced competitors and also exhibit its efficiency and robustness.

\end{abstract}

\section{Introduction}
Heterogeneous information networks (HINs) \cite{sunyizhou2013heteromining,shichuan2016survey} typically include multiple types of nodes and edges, implying the rich semantic information. 
These come together with a lot of real-world data, such as social networks \cite{wang2016structural,hamilton2017sage}, citation networks \cite{atwood2016diffusion,gcn2017kipf} and recommendation systems \cite{zhang2019stargcn,berg2017graph}. 
The complicated heterogeneity and rich semantic information within HINs bring great challenges on heterogeneous graph tasks, such as node classification, link prediction and graph classification.
Recently, the representation learning for heterogeneous graphs \cite{yang2020heterogeneoussurvey} receives a surge of research attention, which presents a great opportunity for analyzing HINs.

To capture both heterogeneity and structural information, heterogeneous graph neural networks (HGNNs) have been proposed and widely used to model HINs in recent years. Existing HGNNs can broadly be categorized into metapath-based models and metapath-free models. 
Generally, metapath-based approaches capture heterogeneity by using the predefined metapaths \cite{han2019wang,magnn2020fu,hetgnn2019zhang,sehgnn2023yang}, while they have to redefine appropriate metapaths to adapt to various heterogeneous graphs. 
To get rid of the dependency on metapaths, metapath-free approaches encode graph heterogeneity by designing additional tailored modules \cite{hgt2020hu,hgb2021lv,hinormer2023mao}.
With the capability of encoding both graph structure and heterogeneity, existing approaches give rise to the performance in a variety of downstream tasks on HINs.
Their success demonstrates that leveraging the heterogeneity of HINs can significantly boost the model's performance. 

\begin{figure}
    \centering
    \includegraphics[width=0.8\linewidth]{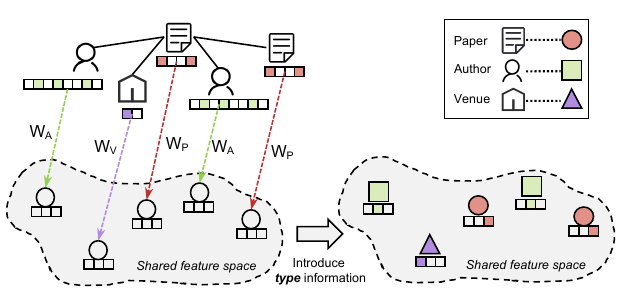}
    \caption{An illustration of the feature processing for a toy citation network. $\text{W}_P$, $\text{W}_A$ and $\text{W}_V$ are type-specific transformation matrices w.r.t. node types.}
    \label{demo_figure}
\end{figure}

Since nodes in different types may have diverse attributes,
most existing metapath-free methods first align node features by projecting them into a shared low-dimensional space \cite{hgb2021lv,hinormer2023mao}.
For example, as illustrated in {Fig}~\ref{demo_figure},
the input feature vectors of papers, authors 
and venues are first mapped into low-dimensional embeddings with same dimension.
Although the low-dimensional 
node embeddings can preserve original feature information and topological information \cite{yang2020heterogeneoussurvey},
node type information is {not retained}. 
This further leads to 
the loss of heterogeneity information in subsequent neighborhood aggregation operation.
Under this condition,
when aggregating information from a node's neighbors,
most existing methods
can only identify 
which neighbors are in the same type,
but they fail to know
the exact types of these neighbors.
Based on the above analysis, the first challenge is to design an encoder that can seamlessly integrate the information of graph heterogeneity, including both node types and edge types,
with node features and graph structure. 

Additionally,
most existing approaches only consider the interactions between nodes while neglecting the latent interactions among different node features \cite{han2019wang,ying2021graphormer,hinormer2023mao}.
Specifically, each convolutional layer can be considered as the one-order interaction between a node and its neighbors.
By stacking multiple convolutional layers,
the high-order information from multi-hop neighboring nodes is captured.
However,
the feature-level high-order information is also useful for label prediction.
For example, in a co-authorship network, attributes of paper nodes include keywords, 
and our target is to predict their research topics.
For a paper, suppose we only consider one keyword like \emph{graph neural networks (GNNs)}, it is difficult to predict whether the paper's label is Information Retrieval (IR) or Artificial Intelligence (AI), as both IR and AI have sub-topics related to GNNs. 
If we consider three keywords \emph{graph neural networks}, \emph{personalized search} and \emph{query recommendation} simultaneously, 
it is easier to classify the paper as IR rather than AI.
This indicates that considering such feature-level interactions can boost the model's capability. 
Therefore, the second challenge is to design an encoder that can capture latent interactions among node features and leverage feature-level high-order information to enhance node embeddings.

To address the challenges, 
in this paper, 
we propose a novel \textbf{Het}erogeneous graph \textbf{C}ascade \textbf{A}ttention \textbf{N}etwork ({\name}). 
\textbf{For the first challenge}, we put forward a type-aware encoder composed of multiple type-aware layers,  
in which learnable type embeddings are explicitly introduced for both nodes and edges. 
The key idea of introducing node type embeddings is to supplement the loss of type information for node embeddings in the low-dimensional space.
To this end,
we first propose to fuse
node feature embeddings with node type embeddings and then obtain the fused node embeddings.
After that,
we use fused node embeddings and edge type embeddings 
to perform attention-based weighted aggregation to learn the type-aware node embeddings.
Owing to type information of both nodes and edges, we can capture heterogeneity, node attributes and graph structure simultaneously at the type-aware encoder.

\textbf{For the second challenge}, inspired by Transformer's outstanding capability of modeling interactions among tokens in the input sequences \cite{transformer2017vaswani,ying2021graphormer}, we propose a dimension-aware encoder to enhance hidden embeddings by learning the feature-level high-order information.
To highlight the importance of node types, 
we also introduce node type embeddings as type encoding which is similar to the positional encoding in Transformer. 
This is intended to distinguish different feature interactions corresponding to node types.
Regarding each dimension (or multiple dimensions) of node embedding as a token, 
we can construct an input sequence for each node.
We then apply multi-head self-attention to these input sequences, allowing each dimension to attend to other dimensions and thereby learn their latent interactions.
The outputs from the dimension-aware encoder are concatenated with those from the type-aware encoder to create the final node representations., i.e., the outputs of one-layer cascade block. 

Overall, the proposed {\name} can be regarded as a cascade model with dual-level awareness, i.e., \textbf{node-level and feature-level} awareness.
On the one hand, each type-aware layer utilizes type embeddings of both nodes and edges, allowing nodes to be aware of their neighborhood’s type information as well as feature information. 
On the other hand, at each dimension-aware layer, we employ multi-head self-attention on the sequences expanded from hidden embeddings and make each dimension to be aware of others, thereby learning the high-order information behind latent feature interactions.
Finally,
the main contributions of this work are summarized as follows:

\begin{itemize}
    \item We present a type-aware encoder to make up for heterogeneous information and capture the node-level high-order information, as well as a dimension-aware encoder for learning the feature-level high-order information.
    \vspace{2pt}
    \item We propose a metapath-free model {\name} built upon the above encoders, which allows us to encode graph heterogeneity in a learnable way and obtain more expressive node representations in an end-to-end manner.
    \vspace{2pt}
    \item We conduct extensive experiments to demonstrate the effectiveness and efficiency of the proposed {\name}.
\end{itemize}

\section{Related Work}
\textbf{Heterogeneous Network Embedding.} 
A large number of graph embedding approaches have been proposed in recent years \cite{perozzi2014deepwalk,grover2016node2vec,tang2015line}, which aim to map nodes or substructures into a low-dimensional space in which the connectivity of the graph can be preserved.
Meanwhile, 
as most of real-world networks are usually composed of various types of nodes and relationships \cite{yang2020heterogeneoussurvey}, researches on heterogeneous network embeddings (HNEs) \cite{shichuan2016survey,wang2022survey} have also received significant attention.
The approaches of HNEs can broadly be  categorized into random walk methods \cite{fu2017hin2vec,dong2017metapath2vec} and first/second-order proximity methods \cite{shi2018easing,tang2015pte}.
\vspace{4pt}

\noindent\textbf{Heterogeneous Graph Neural Networks.}
To capture the rich semantic information contained in heterogeneous graphs, a series of heterogeneous graph neural networks have been proposed in recent years \cite{hgb2021lv,wang2021selfhetero}. According to the way to utilize the graph heterogeneity,  HGNNs are divided into two categories, i.e., metapath-based HGNNs \cite{han2019wang,magnn2020fu,rgcn2018schlichtkrull,sehgnn2023yang} and metapath-free HGNNs \cite{hgt2020hu,hgb2021lv,hinormer2023mao}. 
Typically, metapath-based approaches aggregate messages from type-specific neighboring nodes to generate semantic vectors, then fuse the semantic vectors of different metapaths to output the final node representations. 
Their dependencies on the predefined metapaths make them challenging to apply to complex real-world networks. 
Metapath-free models update node representations by directly employing message passing mechanism, with additional tailored modules to model the heterogeneity, which makes them free from the selection of metapaths. 
But tailored modules tend to make the encoding of heterogeneity separate from node features \cite{hinormer2023mao}, which fails to capture the relations between them. 
We aim to encode both graph heterogeneity and feature information in a unified embedding for all nodes in HINs, 
allowing us to learn more expressive node representations and then improve the model's performance on downstream tasks.


\section{Preliminaries}
\subsection{Heterogeneous Information Network}

A heterogeneous information network  (HIN) can be defined as \begin{math} G=\{\mathcal{V},\mathcal{E}, \mathcal{A}, \mathcal{R} \}\end{math}, where \begin{math} \mathcal{V} \end{math} is the set of nodes as well as \begin{math}\mathcal{E} \end{math} is the set of edges. 
In HIN, each node \begin{math} v \in \mathcal{V} \end{math} has a type \begin{math} \phi (v) \end{math} and each edge \begin{math} e \in\mathcal{E} \end{math} has a type \begin{math} \psi (e) \end{math}. 
The sets of node types and edge types are denoted by \begin{math} \mathcal{A} \end{math} and \begin{math} \mathcal{R}\end{math}, where \begin{math} \mathcal{A} = \{ \phi (v): \forall v \in \mathcal{V} \} \end{math} and \begin{math} \mathcal{R} = \{ \psi (e): \forall e \in\mathcal{E} \} \end{math}, respectively. 
And \begin{math} \phi:\mathcal{V} \rightarrow \mathcal{A} \end{math} is the node type mapping function, \begin{math} \psi:\mathcal{E} \rightarrow \mathcal{R} \end{math} is the edge type mapping function. Typically, a heterogeneous graph is with \begin{math} |\mathcal{A}| + |\mathcal{R}| > 2 \end{math}. Note that, when \begin{math} |\mathcal{A}| = |\mathcal{R}| = 1 \end{math}, the graph degenerates into a homogeneous graph.

\subsection{Graph Neural Networks}
GNNs \cite{gcn2017kipf,gat2018velickkovic} and HGNNs \cite{rgcn2018schlichtkrull,han2019wang,hgb2021lv} commonly rely on the key operation of aggregating neighborhood information in a layer-wise manner, namely the node-level aggregation. In this manner, messages can be recursively passed and transformed from neighboring nodes to the target node. In the $l$-th layer, the representation of node $v$ can be calculated by
\begin{equation}
    \textbf{h}_v^l = \textsc{Aggr}(\textbf{h}_v^{l-1}, \{\textbf{h}_u^{l-1} : u \in \mathcal{N}_v\}; \theta_g^l).
\end{equation}
where $\mathcal{N}_v$ is the neighboring nodes set of node $v$ (or type-specific neighboring nodes for HGNNs), and $\textsc{Aggr}(\cdot ;\theta_g^l)$ denotes the neighborhood aggregation function parameterized by $\theta_g^l$ in the $l$-th layer. 
There are different neighborhood aggregation functions, e.g., mean-pooling aggregation in GCN \cite{gcn2017kipf} and attention-based aggregation in GAT \cite{gat2018velickkovic}.
Since GAT can distinguish different importance of neighboring nodes, we adopt it as the backbone of the proposed type-aware encoder, which will be discussed in next section.

\subsection{Transformer-style Architecture}
\label{3.3}
In the following part, we introduce transformer encoder briefly. The transformer encoder \cite{transformer2017vaswani} is composed of one or multiple transformer blocks, where each transformer block mainly contains a multi-head self-attention (MHSA) module and a feed-forward network (FFN). In natural language processing, the MHSA module, the critical component, aims to receive the semantic correlations among input tokens. 
Regarding each node feature as a token, it can also be generalized to learn the interactions among node features.

Suppose we have an input $\textbf{H} \in \mathbb{R}^{n \times d}$, where $n$ is the length  (or number) of input tokens and $d$ is the hidden dimension. The MHSA firstly projects \textbf{H} to \textbf{Q}, \textbf{K} and \textbf{V} by three linear transformations as 
\begin{equation}
    \textbf{Q} = \textbf{H}\textbf{W}_{\text{q}}, \textbf{K} = \textbf{H}\textbf{W}_{\text{k}}, \textbf{V} = \textbf{H}\textbf{W}_{\text{v}},
\end{equation}
where $\textbf{W}_{\text{q}}, \textbf{W}_{\text{k}} \in \mathbb{R}^{d \times d_{\text{k}}}$ and $\textbf{W}_{\text{v}} \in \mathbb{R}^{d\times d_{\text{v}}}$. Then we calculate the output of MHSA by the scaled dot-product attention mechanism as
\begin{equation}
    \text{MHSA} (\textbf{H}) = \textsc{Softmax} (\frac{\textbf{Q}\textbf{K}^T}{\sqrt{d_{\text{k}}}})\textbf{V},
\end{equation}
where \begin{math} \sqrt{d_{\text{k}}} \end{math} is the scaling factor. 
For simplicity, we use a single-head self-attention module for the description.
Thereafter, the MHSA module is followed by the FFN module which contains two layers of Layer Normalization ($\textsc{LayerNorm}$) and the residual connection \cite{he2016deep}. Then we can obtain the output of $l$-th Transformer block as
\begin{equation}
\begin{split}
    & \textbf{H}^l = \textsc{LayerNorm} (\text{FFN} (\widetilde{\textbf{H}}^l) + \widetilde{\textbf{H}}^l) \\
    & \widetilde{\textbf{H}}^l = \textsc{LayerNorm} (\text{MHSA} (\textbf{H}^{l-1}) + \textbf{H}^{l-1}).
\end{split}
\end{equation}
By stacking $L$ Transformer blocks, we can obtain the final output representation $\textbf{H}^L \in \mathbb{R}^{n \times d}$, which can be used as the input of downstream tasks, such as node classification and link prediction.

\section{The Proposed Model}
\subsection{Overall Architecture}
\begin{figure}[ht]
    \centering
    \includegraphics[width=\linewidth]{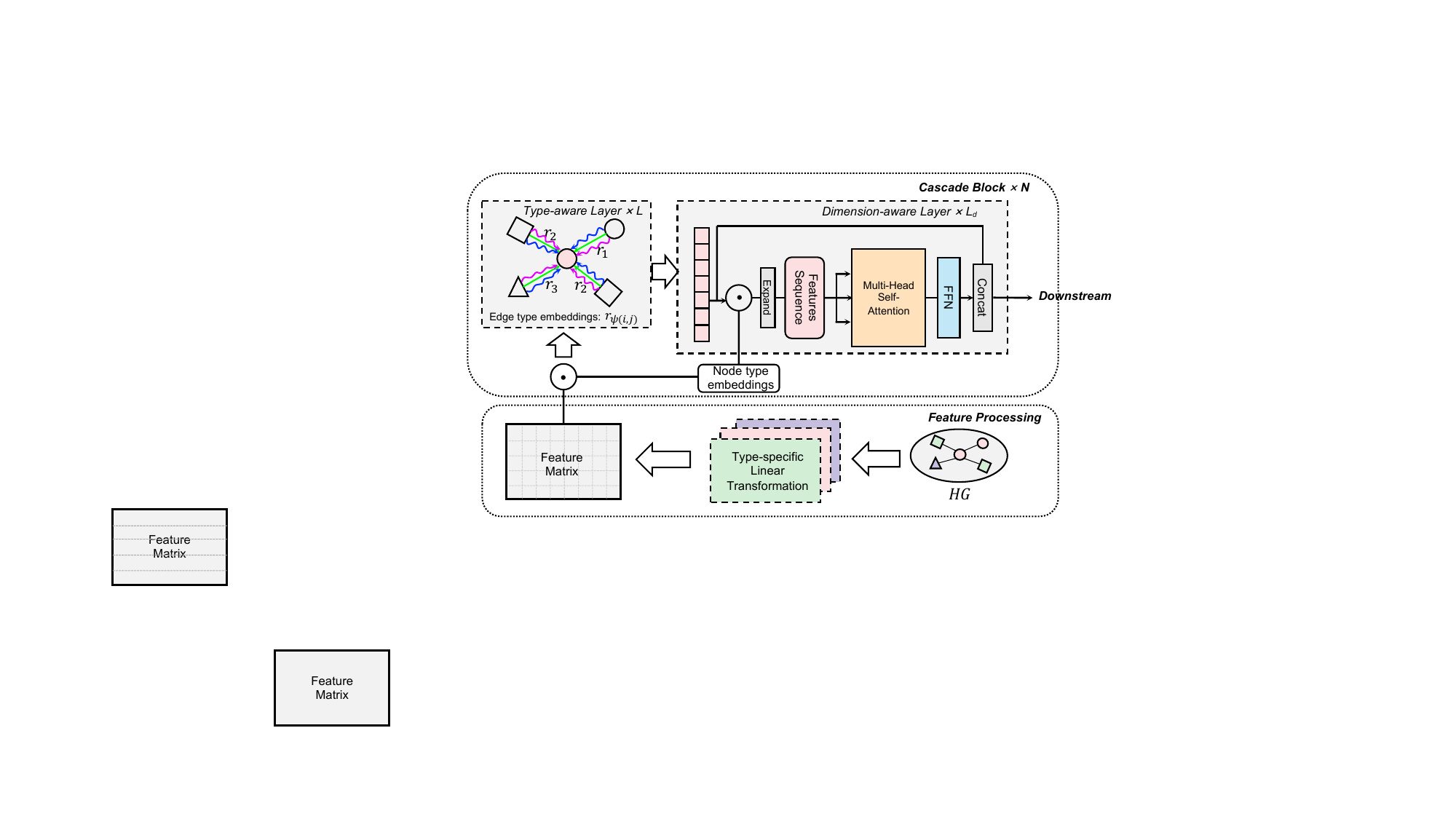}
    \caption{The overall framework of {\name}. Each cascade block consists of $L$ type-aware layers and $L_d$ dimension-aware layers.}
    \label{architecture}
\end{figure}

The overall framework of {\name} is illustrated in Fig~\ref{architecture}. 
Given a heterogeneous graph \textit{HG}, we first adopt type-specific linear transformations to project nodes with different feature spaces into a shared feature space.
Then, the aligned embeddings are employed as initial node feature matrix and are fed into the type-aware encoder, where each node can simultaneously perceive heterogeneity information, feature information and structural information within its neighborhood.
After multiple type-aware layers, hidden node embeddings are passed to the dimension-aware encoder, where latent feature interactions will be modeled through multi-head self-attention mechanism.
Afterward, we concatenate the outputs from the type-aware encoder and the dimension-aware encoder to construct the updated node embeddings, which are also referred to the outputs of each cascade block.
{\name} typically includes $N$ cascade blocks. 
Finally, we perform downstream tasks based on the normalized final node representations. 
In the following parts, we will illustrate the details of the type-aware encoder and the dimension-aware encoder, respectively.
\subsection{Type-aware Encoder}
In the type-aware encoder,
we first introduce learnable type embeddings for both nodes and edges, and integrate feature embeddings and type embeddings as a whole.
Formally, we first initialize a node type matrix denoted by $\mathbf{M} \in \mathbb{R}^{|\mathcal{A}|\times d_t}$, where $|\mathcal{A}|$ is the number of node types. For each node $v_i$, its node type embedding $\mathbf{t}_i \in \mathbb{R}^{d_t}$ is derived by $\mathbf{t}_i = \mathbf{M}[\phi (v_i),:]$. 
Then we can obtain type embeddings of all nodes represented as $\mathbf{T} \in \mathbb{R}^{n \times d_t}$. 
As node features of different types on HINs usually exist in different feature spaces, we project them into a shared feature space before passing them to the type-aware encoder. Formally, the feature processing is denoted as
\begin{equation}
    \mathbf{h}_i = \mathbf{W}_{\phi (v_i)}\mathbf{x}_i + \mathbf{b}_{\phi (v_i)},
\label{linear transformation}
\end{equation}
where $\mathbf{W}_{\phi (v_i)} \in \mathbb{R}^{d\times d_x}$ is the learnable parameter matrix corresponding to node type $\phi (v_i)$ and $\mathbf{b}_{\phi (v_i)} \in \mathbb{R}^{d}$ is an optional bias term. Then we can obtain node feature embeddings denoted by $\mathbf{H} \in \mathbb{R}^{n\times d}$.
After that, to comprehensively supplement node type information, 
we present a combination function to integrate node feature embeddings with node type embeddings as
\begin{equation}
    \mathbf{\widetilde{H}} = \textsc{Combine}(\mathbf{H}, \mathbf{T}),
\label{fusion}
\end{equation}
where \textsc{Combine} (·) can be any operator function, such as learnable functions or non-parametric functions.
In practice, we simply implement it with Hadamard product which is an element-wise operation.
Based on \cite{hgb2021lv}, we then extend attention mechanism with integrated node embeddings that contain the node type information. In this way, each type-aware layer calculates the attention coefficient between node $v_i$ and node $v_j$ as follows (layer marker $(l)$ is omitted for simplicity)
\begin{equation}
    \alpha_{ij} = \frac{
    \text{exp}\left ( \sigma \left ( \mathbf{a}^T[\mathbf{W}\mathbf{\tilde{h}}_i || \mathbf{W}\mathbf{\tilde{h}}_j || \mathbf{W_r}\mathbf{r}_{\psi (v_i, v_j)}]   \right) \right)
    }{
    \sum_{k \in \mathcal{N}_i}{\text{exp}\left ( \sigma \left ( \mathbf{a}^T[\mathbf{W}\mathbf{\tilde{h}}_i || \mathbf{W}\mathbf{\tilde{h}}_k || \mathbf{W_r}\mathbf{r}_{\psi (i,k)}]   \right) \right)}
    },
\end{equation}
where $\mathbf{r}_{\psi (v_i, v_j)} \in \mathbb{R}^{d_r}$ is learnable edge type embedding w.r.t. the type of edge between node $v_i$ and node $v_j$,
$\mathbf{W}$ and $\mathbf{W}_r$ are learnable matrices, and $\sigma$ is the $\text{LeakyReLU}$ activation function. 
To stabilize training process and improve the performance, inspired by \cite{he2016deep,he2020realformer}, we employ residual connection on attention coefficients as
\begin{equation}
    \hat{\alpha}_{ij}^{ (l)} =  (1-\beta) \alpha_{ij}^{ (l)} + \beta \hat{\alpha}_{ij}^{ (l-1)},
\end{equation} where $\beta \in [0,1]$ denotes attention residual weight.
Once obtained, the normalized attention coefficients are used to update the hidden node embedding $\mathbf{h}'_i$ for each node $v_i \in \mathcal{V}$ as 
\begin{equation}
    \mathbf{h}'_i = \sigma \left ( \sum_{j\in \mathcal{N}_i}\hat{\alpha}_{ij}\mathbf{W}\mathbf{\tilde{h}}_j\right).
\end{equation}
To enhance model's capacity and stabilize the learning process, we implement a multi-head attention mechanism by averaging
\begin{equation}
    \mathbf{h}'_i = \sigma \left ( \frac{1}{K} \sum_{k=1}^{K}\sum_{j\in \mathcal{N}_i}\hat{\alpha}^{k}_{ij}\mathbf{W}^{k}\mathbf{\tilde{h}}_j\right),
\label{gat output}
\end{equation}
where $K$ is the number of heads. 
Overall, with the type-aware encoder, 
hidden node embeddings can seamlessly fuse the information of graph heterogeneity, node feature and graph structure and have more powerful expressive capabilities.

\subsection{Dimension-aware Encoder}
The success of Transformer has demonstrated its outstanding capability of learning interactions among the tokens in a sequence. 
Motivated by this, 
we propose a dimension-aware encoder with transformer architecture for capturing the feature-level high-order information,
which can further enhance the expressive capability of node embeddings.

After acquiring hidden embeddings $\mathbf{H'}\in \mathbb{R}^{n\times d}$ from $L$ type-aware layers, 
the dimension-aware encoder constructs the input sequence for each node to adapt to transformer architecture.
Specifically, for each node $v \in \mathcal{V}$, we expand its hidden embedding $\mathbf{h}'_v \in \mathbb{R}^d$ to a sequence $\mathbf{\hat{h}}'_v \in \mathbb{R}^{d \times 1}$, treating each dimension (or multiple dimensions) as a token represented by a one-dimensional (or multiple dimensional) vector.
Then we perform multi-head self-attention on each input sequence to learn interactions among the tokens within it.

Besides, to learn distinct feature interaction patterns for different node types, inspired by the positional encoding in Transformer, we introduce node type embeddings $\mathbf{T} \in \mathbb{R}^{n \times d_t}$ as type encoding and combine them with the hidden node embeddings $\mathbf{H}'$ before performing attention mechanism. 
We denote this step as
\begin{equation}
\begin{split}
     &\mathbf{\hat{H}} = \textsc{Combine} (\mathbf{H'}, \mathbf{T}) \\
     &\mathbf{\hat{H}'} = \textsc{Expand} (\mathbf{\hat{H}}) 
\end{split}
\end{equation}
where $\mathbf{\hat{H}'} \in \mathbb{R}^{n\times d \times 1}$ denotes the constructed sequences of all nodes, illustrated in upper right of Fig~\ref{architecture}. Similar to the type-aware encoder, the \textsc{Combine} (·) is also implemented with Hadamard product by simply setting $d_t = d$, so that the shape of  $\mathbf{\hat{H}}$ remains consistent with $\mathbf{H'}$.
With type encoding, the dimension-aware encoder can distinguish various node types and learn unique interaction patterns for them. 
Then, we perform multi-head self-attention on the input sequences $\mathbf{\hat{H}'}$ and obtain the outputs of each dimension-aware layer as follows
\begin{equation}
    \mathbf{\overline{H}}= \textsc{Mhsa} (\mathbf{\hat{H}'}),
\end{equation}
where $\textsc{Mhsa}(\cdot)$ denotes multi-head self-attention (refer to Section~\ref{3.3}) and
$\mathbf{\overline{H}} \in \mathbb{R}^{n\times d}$ is the node representations containing rich feature-level information. 
Finally, we concatenate the outputs of dimension-aware encoder $\mathbf{\overline{H}}$ and the type-aware encoder $\mathbf{H'}$ to construct the final node representations as
\begin{equation}
    \mathbf{H}_{\text{f}} = \mathbf{H'} \parallel \mathbf{\overline{H}},
\end{equation}
where 
$\mathbf{H}_{\text{f}} \in \mathbb{R}^{n \times 2d}$ is the output of one-layer cascade block. For simplicity, we only illustrate one layer of cascade block. 
After one or multiple cascade blocks, we can obtain the enhanced node representations with more expressive capabilities and use them for various downstream tasks. 

\subsection{Time Complexity Analysis}
In this subsction, we give the time complexity analysis of the proposed components in {\name}.
Let $|{\mathcal{V}}|$ and $|{\mathcal{E}}|$ are the number of nodes and edges.
$d$ is the dimension of both node feature embeddings and node type embeddings, and $d_r$ is the dimension of edge embeddings.
For each type-aware layer, the time complexity of a single attention head can be expressed as $O(|{\mathcal{V}}|\times d^2+|{\mathcal{E}}| \times {d_r}^2 + |{\mathcal{E}}| \times (2d + d_r))$. For each dimension-aware layer, the time complexity of a single attention head can be expressed as $O(|{\mathcal{V}}|\times d^2)$. Thus, overall time complexity of {\name} is linear to both the number of nodes $|{\mathcal{V}}|$ and the number of edges $|{\mathcal{E}}|$. The efficiency studies of our model are shown in Fig~\ref{Efficiency study}.

\section{Experiments}
We evaluate {\name} by conducting extensive experiments on node classification and link prediction, and compare various competitive approaches, including plain homogeneous GNNs, metapath-based HGNNs and metapath-free HGNNs. 
In addition, to further investigate the superiority of our model, we comprehensively conduct three studies including an ablation study, an efficiency study and a parameter study.
The source code and datasets are available at https://github.com/zzyzeyuan/HetCAN.

\subsection{Experimental Setups}
\textbf{Datasets}. For node classification, we test our model on five public datasets. Specifically, DBLP, IMDB, ACM and Freebase are from \cite{hgb2021lv}, 
and OGB-MAG is from \cite{hu2020ogb}. For link prediction, we test our model on three public datasets from \cite{hgb2021lv}. 
Heterogeneous Graph Benchmark standardizes the process pipeline for fair comparison, so we follow their pipelines to conduct experiments. 
For datasets without node features, we assign them one-hot or all-one vector features to denote their existence.
The statistics of all datasets are summarized in Table~\ref{Statistics of all datasets}.  

\begin{table}
    \centering
    \caption{Statistics of all datasets.}
    \label{Statistics of all datasets}
    \renewcommand{\arraystretch}{1} 
    \setlength{\tabcolsep}{2pt}
    \resizebox{0.8\linewidth}{!}{
    \begin{tabular}{c|rcrccc}
        \toprule
        \multirow{2}{*}{Datasets} & \multirow{2}{*}{\#{Nodes}} & \multicolumn{1}{c}{\#Node} & \multirow{2}{*}{\#Edges} & \multicolumn{1}{c}{\#Edge} & \multirow{2}{*}{Target} & \multirow{2}{*}{\#Classes} \\
        & & \multicolumn{1}{c}{Types} &  & \multicolumn{1}{c}{Types} & &  \\
        \midrule
        DBLP & 26,128& 4&239,566&6 & author&4\\
        IMDB & 21,420& 4 & 86,642&6&movie&5\\
        ACM & 10,942 & 4 & 547,872&8&paper&3\\
        Freebase & 180,098& 8 & 1,057,688&36&book&7\\
        OGB-MAG & 1,939,743 & 4 & 21,111,007 & 4& paper&349 \\
        \midrule
        Amazon & 10,099& 1& 148,659&2 & product-product& -\\
        LastFM &20,612 & 3& 141,521&3 & user-artist& -\\
        PubMed &63,109 & 4& 244,986& 10& disease-disease&- \\
    \bottomrule
    \end{tabular}}    
\end{table}

\noindent\textbf{Baselines}. To comprehensively evaluate our proposed model against the state-of-the-art methods, we select a collection of baselines, including \textit{basic models}  (GCN \cite{gcn2017kipf}, GAT \cite{gat2018velickkovic}, Transformer \cite{transformer2017vaswani}), 
\textit{metapath-based models} 
(RGCN \cite{rgcn2018schlichtkrull}, 
HAN \cite{han2019wang}, 
HetGNN \cite{hetgnn2019zhang}, 
MAGNN \cite{magnn2020fu}, 
SeHGNN \cite{sehgnn2023yang}) and 
\textit{metapath-free models}
HGT \cite{hgt2020hu}, 
Simple-HGN \cite{hgb2021lv}, 
HINormer \cite{hinormer2023mao}). 
Specifically, as all baselines do not utilize extra label embeddings, we report the results of SeHGNN without the utilization of extra label embeddings. 
\vspace{4pt}

\noindent\textbf{Settings}. 
Regarding datasets from HGB, we follow the split proportion of 24:6:70 for the training, validation and test sets, respectively. 
Regarding OGB-MAG dataset, we propose to
train on papers published until 2017, validate on those published in 2018, and test on those published
since 2019. 
We evaluate classification performance of all baselines with Micro-F1 and Marcro-F1 for HGB datasets and accuracy for OGB-MAG dataset. 
Following HGB, we use ROC-AUC (area under the ROC curve) and MRR (mean reciprocal rank) metrics to evaluate link prediction performance.
Since our experimental setup is consistent with HGB and OGB, we directly borrow the results reported in HGB and OGB leaderboard for comparison.
For those results that are not available in HGB or OGB, 
we conduct experiments based on original experimental setups.


\begin{table*}
    \caption{Experiment results on four HGB datasets. The best result is \textbf{bolded} and the runner-up is \underline{underlined}. The error bar (±) denotes the standard deviation of the results over five runs. ``-'' denotes the models run out of memory.}
    \renewcommand{\arraystretch}{1.2} 
    \label{BIG table}
    \resizebox{\linewidth}{!}{
    \begin{tabular}{l cc cc cc cc}
    \toprule
    \multirow{2}{*}{Methods}&\multicolumn{2}{c}{DBLP} &\multicolumn{2}{c}{IMDB}&\multicolumn{2}{c}{ACM}&\multicolumn{2}{c}{Freebase} \\
    & Micro-F1 & Macro-F1& Micro-F1& Macro-F1& Micro-F1& Macro-F1& Micro-F1& Macro-F1\\
    \midrule
    GCN & 91.47±0.34&90.84±0.32 & 64.82±0.64& 57.88±1.18& 92.12±0.23&92.17±0.24 &60.23±0.92 & 27.84±3.13   \\
    GAT& 93.39±0.30 & 93.83±0.27& 64.86±0.43&58.94±1.35 &92.19±0.93 &92.26±0.94&65.26±0.80 & 40.74±2.58  \\
    Transformer & 93.99±0.11& 93.48±0.12&66.29±0.69 &62.79±0.65 & 92.32±0.36& 92.55±0.34& 65.74±0.78& 47.06±2.42 \\
    \midrule
    HGT & 93.49±0.25&93.01±0.23 &67.20±0.57 &63.00±1.19 & 91.00±0.76& 91.12±0.76&60.51±1.16 & 29.28±2.52\\
    Simple-HGN &94.46±0.22 &94.01±0.24 & 67.36±0.57& 63.53±1.36& 93.35±0.45&93.42±0.44& \underline{66.29±0.45}& 47.72±1.48 \\ 
    HINormer & 94.94±0.21& 94.57±0.23 &67.83±0.34 & 64.65±0.53& 92.13±0.32&92.20±0.34 & 66.08±0.74& 49.37±2.08 \\
    \midrule
    RGCN & 92.07±0.50& 91.52±0.50& 62.05±0.15&58.85±0.26 & 91.41±0.75&91.55±0.74 & 58.33±1.57& 46.78±0.77 \\
    HAN & 92.05±0.62& 91.67±0.49&64.63±0.58 &57.74±0.96 &90.79±0.43 &90.89±0.43 & 54.77±1.40& 21.31±1.68\\
    MAGNN & 93.76±0.45&93.28±0.51 & 64.67±1.67& 56.49±3.20& 90.88±0.64&90.77±0.65 & -& -  \\
    SeHGNN &\underline{95.24±0.13} &\underline{94.86±0.14} &\underline{68.21±0.32} & \underline{66.63±0.34}&\underline{93.87±0.50} & \underline{93.95±0.48}& 63.41±0.47&\underline{50.71±0.44}\\
    \midrule
    
    {\name} &\textbf{95.78±0.28}&\textbf{95.45±0.23} & \textbf{69.50±0.34}& \textbf{67.23±0.28} &\textbf{94.35±0.35} & \textbf{94.47±0.36}& \textbf{66.79±0.52}& \textbf{51.48±0.63} \\
    \bottomrule
  \end{tabular}}
\end{table*}

\subsection{Node Classification}
Tables \ref{BIG table} and \ref{ogb results} summarize experimental results on node classification over five runs. 
From the tables, we observe that:
\vspace{4pt}

\noindent(1) 
The plain models, i.e., GCN, GAT and Transformer, perform well on all datasets when using proper inputs from HGB, indicating that preprocessing for input node features has great impact on model performance. 
\vspace{4pt}

\noindent(2) Compared to the vanilla models mentioned earlier, SeHGNN and HINormer demonstrate superior performance, with SeHGNN being the best among metapath-based models and HINormer excelling among metapath-free models. 
By using the predefined metapaths, SeHGNN exploits semantic information to boost model performance.
HINormer samples a fixed-length sequence for each node and designs an additional heterogeneous relation encoder, which enlarges the receptive field for each node and also models the heterogeneity.
\vspace{4pt}

\noindent(3) 
{\name} achieves superior results across all HGB datasets, demonstrating its ability to generalize to datasets with varying degrees of heterogeneity. 
We attribute the generalization ability of our model to the Cascade Block, which allows us to simultaneously learn node-level and feature-level information. This enables the node representations to have more powerful expressive capabilities, thereby boosting both node classification and link prediction tasks.
\begin{table}[ht]
    \caption{Experiment results on the large-scale dataset OGB-MAG. $*$ denotes metapath-free models. The best result is \textbf{bolded} and the runner-up are \underline{underlined}.}
    \label{ogb results}
    \centering
    \renewcommand{\arraystretch}{1} 
    \resizebox{0.5\linewidth}{!}{
    \begin{tabular}{l c c}
        \toprule
        Methods & Validation accuracy&  Test accuracy\\
        \midrule
        RGCN & 48.35±0.36& 47.37±0.48  \\
        HGT* &49.89±0.47 &49.27±0.61  \\
        NARS &51.85±0.08 & 50.88±0.12\\
        SAGN* &52.25±0.30& 51.17±0.32\\
        GAMLP* & 53.23±0.23& 51.63±0.22\\
        LEGNN* & 54.43±0.09& 52.76±0.14\\
        SeHGNN & 55.95±0.11& \textbf{53.99±0.18}\\
        \midrule
        {\name}* & 54.76±0.18& \underline{53.79±0.17}\\
        \bottomrule
    \end{tabular}}
\end{table}

\vspace{4pt}
\noindent(4) Based on the results from the large-scale dataset OGB-MAG (see Table~\ref{ogb results}), we can observe that {\name} outperforms all metapath-free competitors.
This indicate that our method has further narrowed the gap between metapath-free models and metapath-based models on large-scale dataset. 
In addition, we further conduct efficiency studies on three datasets shown in Fig~\ref{Efficiency study}, which demonstrates that our method is faster than SeHGNN, the winner of OGB-MAG dataset. 
Particularly, shown in Fig~\ref{Efficiency study}(c), the convergence speed of our model is much faster than SeHGNN in the scenario with a large number of edge types (39 edge types in Freebase). This is because metapath-based methods require aggregating information through metapaths, and this inherent property results in slower training and convergence speeds.

\subsection{Link Prediction}
Table~\ref{link prediction} summarizes the results on the downstream link prediction task over five runs. Based on this table, we observe that:
\vspace{3pt}

\noindent(1)
Our method {\name} consistently outperforms all advanced methods over both ROC-AUC and MRR metrics. 
Particularly, we achieve significant improvements on the Amazon and LastFM datasets.
This indicates that our method can learn more expressive node representations with such a cascade structure, while also giving rise to the model's performance on the link prediction task.
\vspace{3pt}

\noindent(2) 
Compared to Simple-HGN, the runner-up on link prediction, our method achieves better performance. 
Our method introduces both the node-level and the feature-level high-order information through the cascade block, while Simple-HGN only uses learnable type embeddings to compensate for graph heterogeneity and ignores the feature high-order interactions.

\begin{table}
    \caption{Experiment results on link prediction. The best result is \textbf{bolded} and the runner-up is \underline{underlined}. The error bar (±) denotes the standard deviation of the results over five runs. ``-'' denotes the results are not available due to lack of metapaths on those datasets.}
    \vspace{3pt}
    \label{link prediction}
    \renewcommand{\arraystretch}{1} 
    \centering
    \resizebox{0.95\linewidth}{!}{
    \begin{tabular}{lcccc cc}
    \toprule
    \multirow{2}{*}{Methods}&\multicolumn{2}{c}{Amazon} &\multicolumn{2}{c}{LastFM}&\multicolumn{2}{c}{PubMed}\\
    & ROC-AUC & MRR & ROC-AUC & MRR & ROC-AUC & MRR\\
    \midrule
    GCN &92.84±0.34 &\underline{97.05±0.12} &59.17±0.31 & 79.38±0.65 &80.48±0.81 & 90.99±0.56\\
    GAT & 91.65±0.80& 96.58±0.26& 58.56±0.66& 77.04±2.11& 78.05±1.77&90.02±0.52\\
    \midrule
    RGCN &86.34±0.28 &93.92±0.16 & 57.21±0.09&77.68±0.17& 78.29±0.18& 90.26±0.24\\
    GATNE &77.39±0.50 &92.04±0.36 &66.87±0.16 & 85.93±0.63& 63.39±0.65&80.05±0.22\\
    HetGNN & 77.74±0.24&91.79±0.03 &62.09±0.01 &83.56±0.14&73.63±0.01 &84.00±0.04\\
    \midrule
    HGT & 88.26±2.06& 93.87±0.65& 54.99±0.28& 74.96±1.46& 80.12±0.93& 90.85±0.33\\
    Simple-HGN & \underline{93.40±0.62}&96.94±0.29 & \underline{67.59±0.23}& \underline{90.81±0.32}& \underline{83.39±0.39}&\underline{92.07±0.26}\\
    \midrule
    {\name} & \textbf{95.60±0.32}&\textbf{98.14±0.28} &\textbf{67.92±0.17}&\textbf{91.78±0.40}&\textbf{83.94±0.30} &  \textbf{92.48±0.24}\\
    \bottomrule
    \end{tabular}}
\end{table}

\begin{figure}[ht]
    \begin{minipage}{0.5\linewidth}
        \centering
        \includegraphics[width=\linewidth]{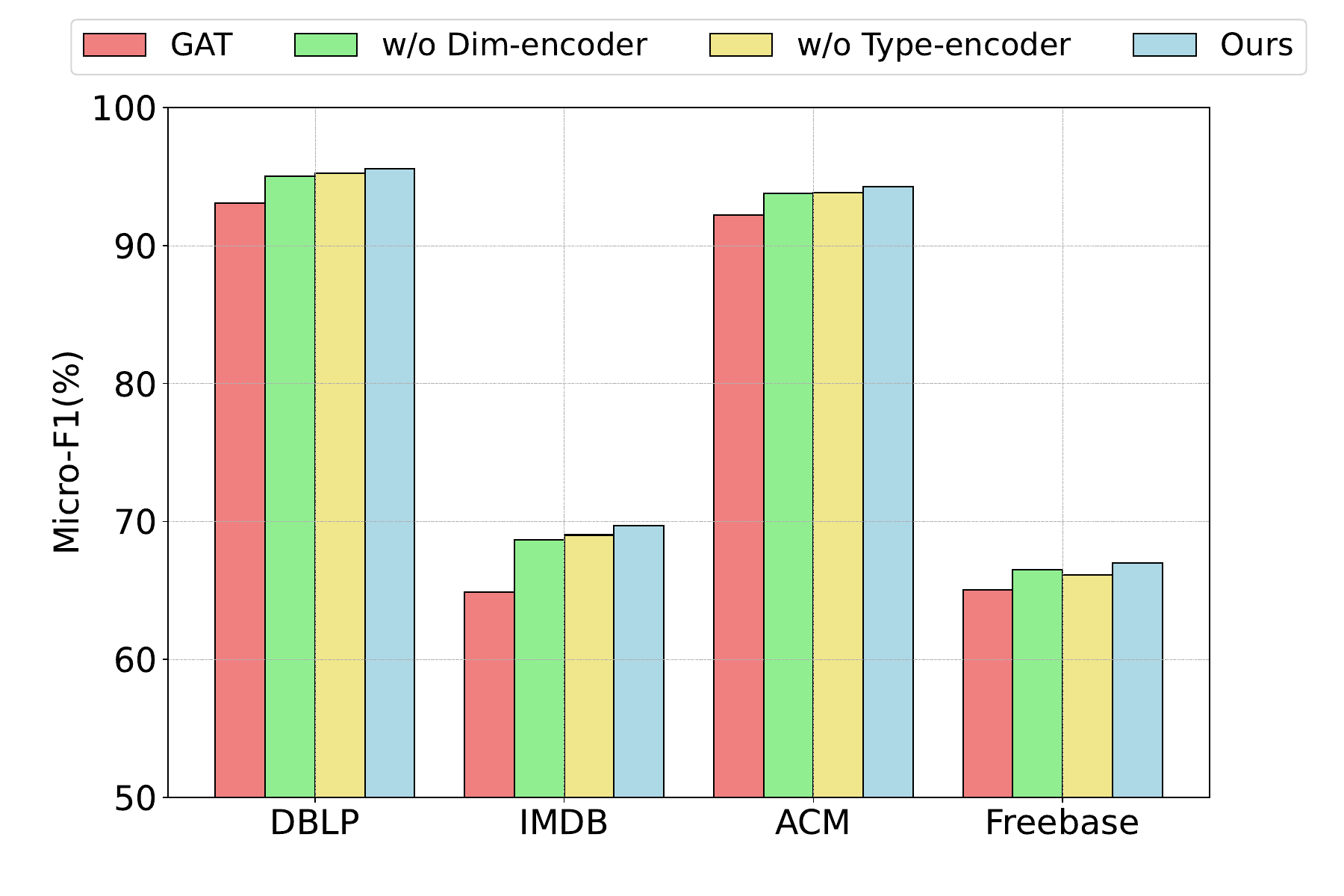}
    \end{minipage}
    \begin{minipage}{0.5\linewidth}
        \centering
        \includegraphics[width=\linewidth]{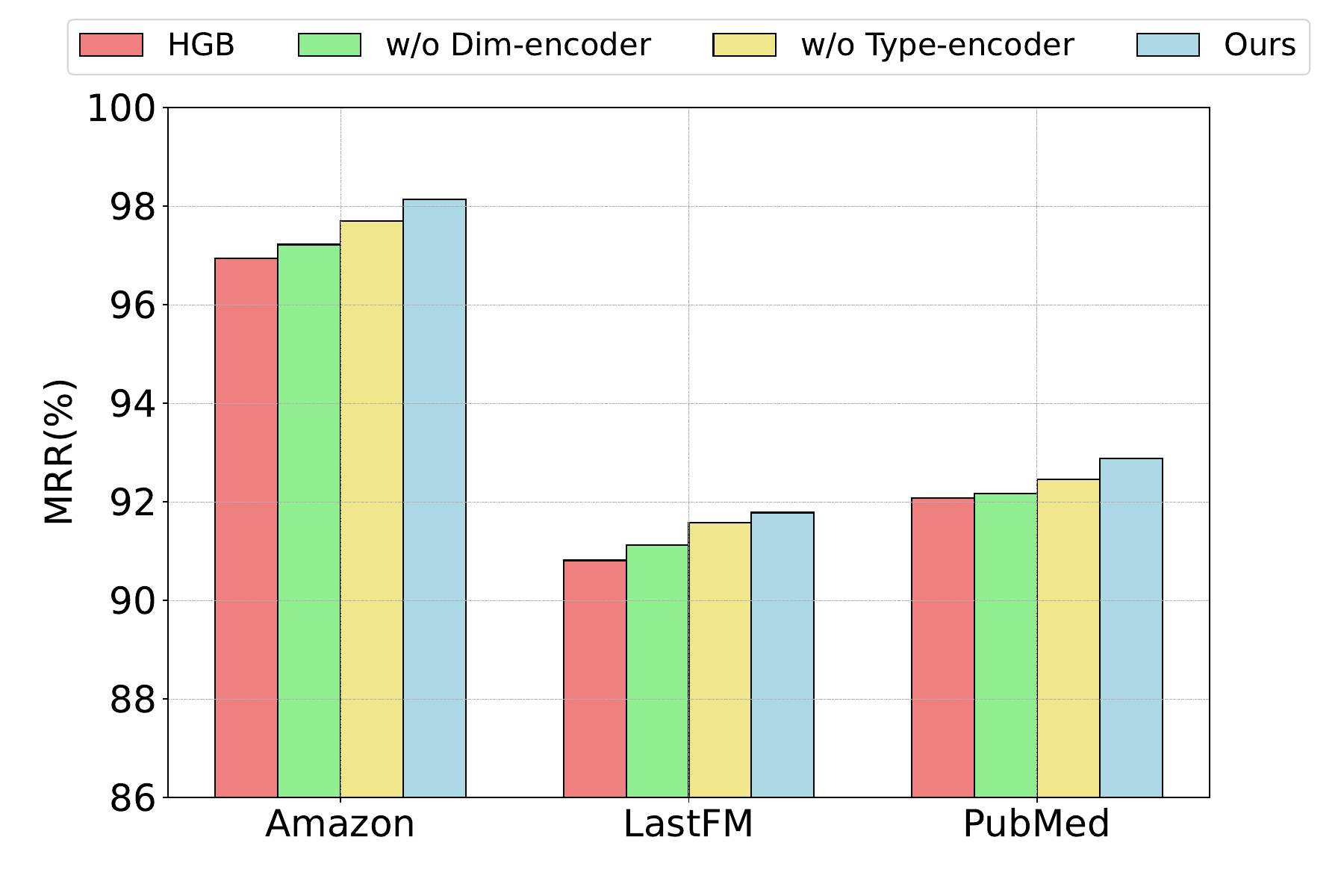}
    \end{minipage}
    \caption{Ablation studies.}
    \label{ablation_fig}
\end{figure}

\subsection{Model Analysis}
\noindent\textbf{Ablation studies.} To validate effectiveness of the proposed components, we conduct ablation studies on four datasets by comparing with two variants of {\name}:  (1) we remove node type embeddings and replace it with all-one vectors, which is denoted by \textit{w/o Type-encoder};  (2) we remove the dimension-aware encoder, which is denoted by \textit{w/o Dim-encoder}. We report the results of ablation studies in Fig~\ref{ablation_fig} and give the following observations.

Firstly, without 
the type-aware encoder, {\name} fails to consider node type when performing attention mechanism, resulting in degradation of classification performance. 
Compared to other datasets, the absence of node type embeddings has more prominent impacts  on Freebase that has more node types, indicating that introducing  learnable node type embeddings explicitly can make up for type information and 
benefit the model's performance. 
From another perspective, improvements on some datasets are not as significant as that on Freebase, thus how to further exploit the underlying semantic relations between nodes is still a promising direction and we will investigate it in future work.

Secondly, without dimension-aware encoder, {\name} fails to capture latent feature interactions, resulting in a significant reduction of performance.
We also notice that the absence of dimension-aware encoder has a more significant impact on Macro-F1 scores than Micro-F1. 
Especially on Freebase, the Macro-F1 is reduced by 2.88\% and has a larger standard deviation, which indicates that   
dimension-aware encoder can benefit the robustness of our model.

\begin{figure}
    \centering
    \begin{minipage}{0.33\linewidth}
        \includegraphics[width=\linewidth]{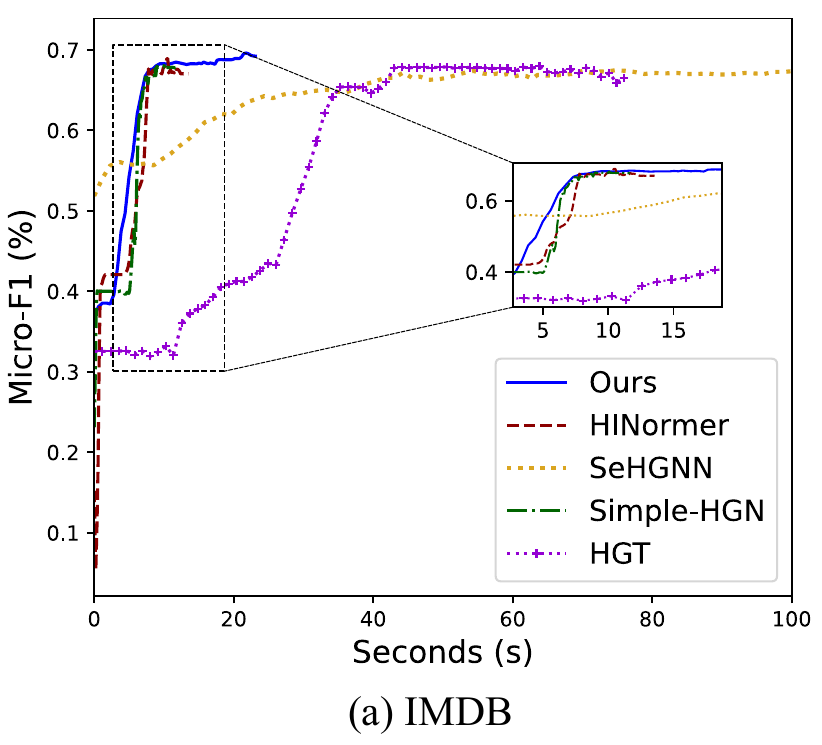}
    \end{minipage}
    \hspace{-2mm}
    \begin{minipage}{0.33\linewidth}
        \includegraphics[width=\linewidth]{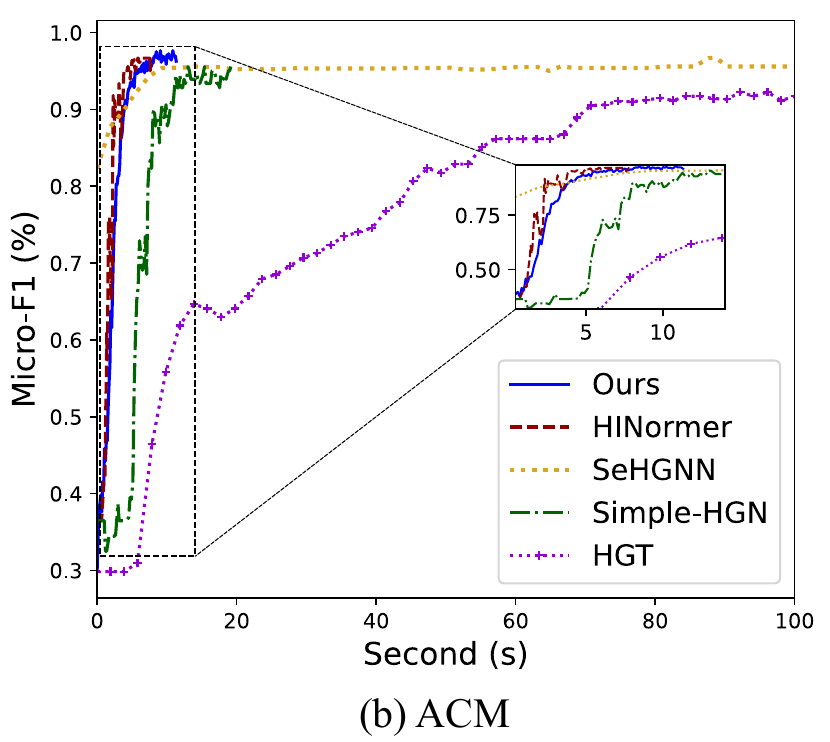}
    \end{minipage}
    \hspace{-2mm}
    \begin{minipage}{0.33\linewidth}
        \includegraphics[width=\linewidth]{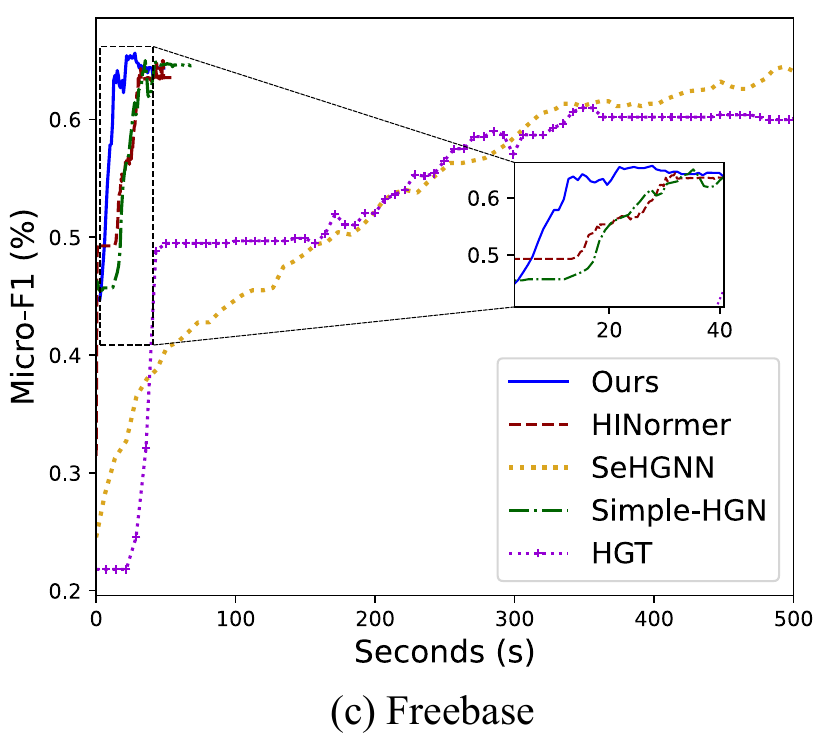}
    \end{minipage}
    \hspace{-2mm}
    \caption{Efficiency study: x-axis shows the training time and y-axis is the Micro-F1 score on the validation set.}
    \label{Efficiency study}
\end{figure}

\vspace{3pt}
\noindent \textbf{Efficiency studies}. 
\label{5.4}
To assess the efficiency of {\name}, we compared the training times of several advanced methods in the same experimental environment, using the hyper-parameters corresponding to optimal performance for each method.
The results are illustrated in Fig~\ref{Efficiency study}.
Specifically, on IMDB, {\name} converges in around 10 seconds, while SeHGNN and HGT take more than 30 seconds. This indicates that our model is as efficient as other metapath-free methods and significantly faster than SeHGNN and HGT.
On Freebase, {\name} achieves the optimal performance around 20 seconds, while HINormer and Simple-HGN approach their optimal state around 40 seconds. 
This also demonstrates the efficiency and robustness of {\name} on information networks with a greater variety of node types and edges.
Surprisingly, we found that SeHGNN takes approximately 500 seconds, 20 times of our model, to converge to the optimal state. 
This demonstrates the superiority and flexibility of being free from the predefined metapaths.

\begin{figure}
    \centering
    \includegraphics[width=0.5\linewidth]{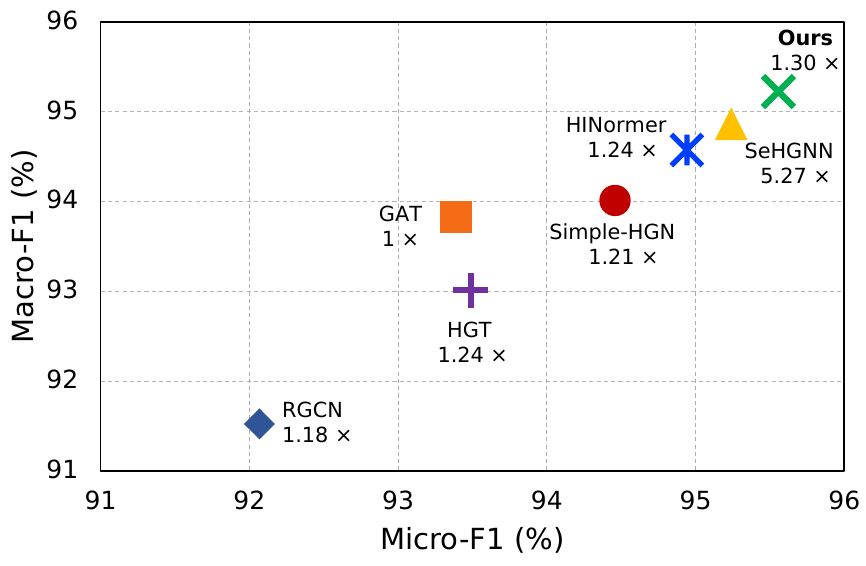}
    \caption{Parameters comparison. The numbers below the model names represent the ratio of the total number of parameters relative to GAT. For example, ``1.24'' below HGT means its total parameters are 1.24 times that of GAT.}
    \label{rank}
\end{figure}

\noindent \textbf{Parameter studies}.
We experiment on DBLP (Fig~\ref{rank}) to statistically compare {\name}'s total parameter count with that of other competitors. 
We use the hyper-parameters corresponding to the optimal performance of these models.
We observe that SeHGNN achieves its peak performance with a large hidden size (512), which leads to a slower convergence speed. 
In contrast, {\name} achieves state-of-the-art performance by introducing an affordable number of parameters, ensuring both efficiency and effectiveness.

\begin{figure}
    \begin{subfigure}{0.33\linewidth}
        \centering
        \includegraphics[width=\linewidth]{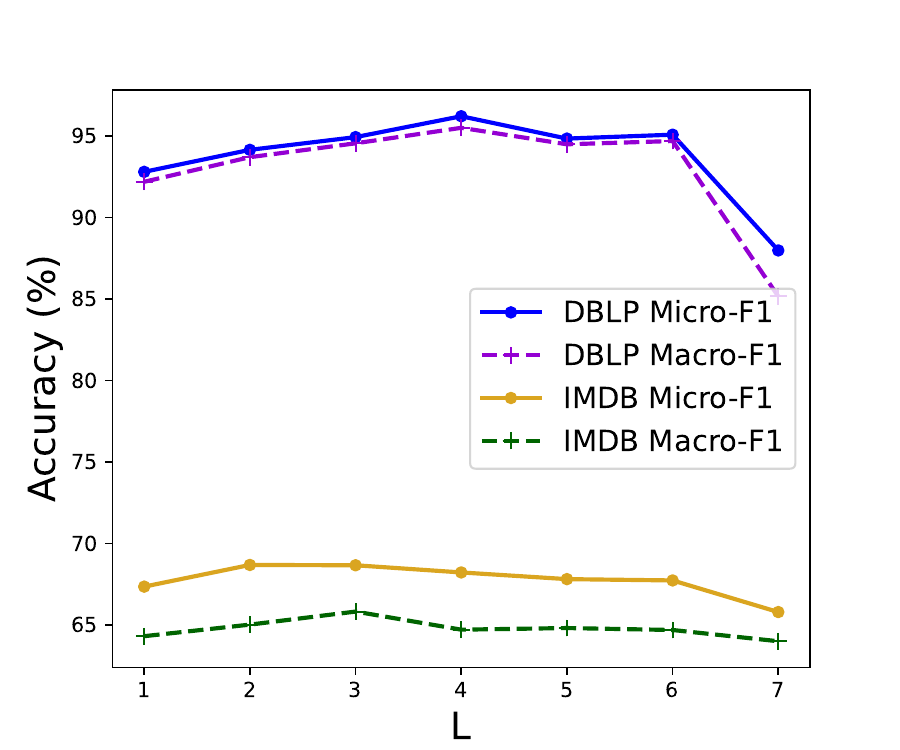}
    \end{subfigure}
    \hspace{-1mm}
    \begin{subfigure}{0.33\linewidth}
        \centering
        \includegraphics[width=\linewidth]{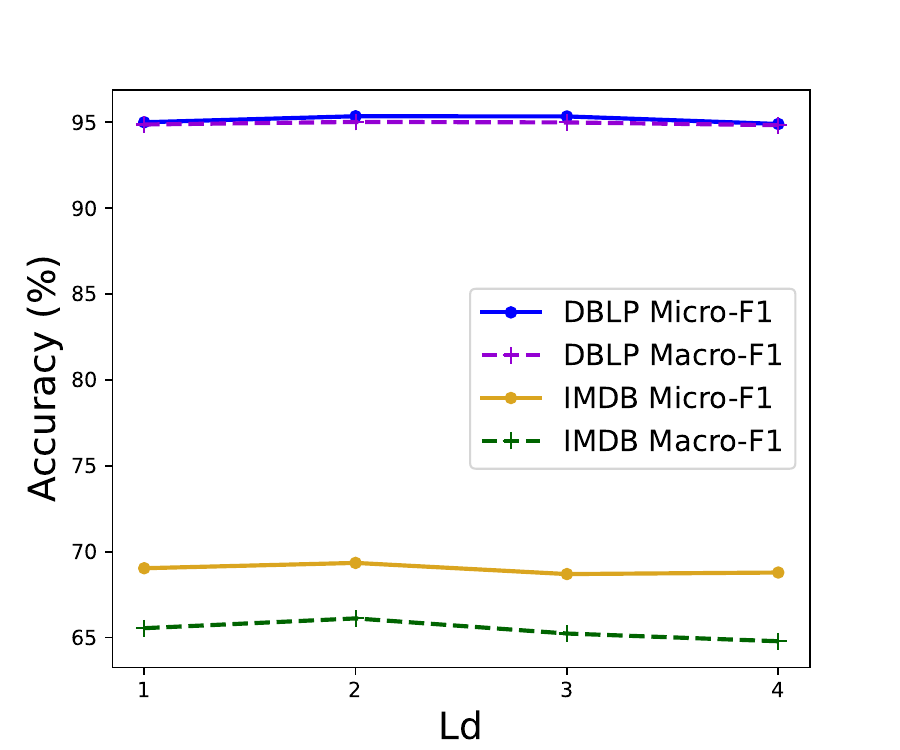}
    \end{subfigure}
    \hspace{-1mm}
    \begin{subfigure}{0.33\linewidth}
        \centering
        \includegraphics[width=\linewidth]{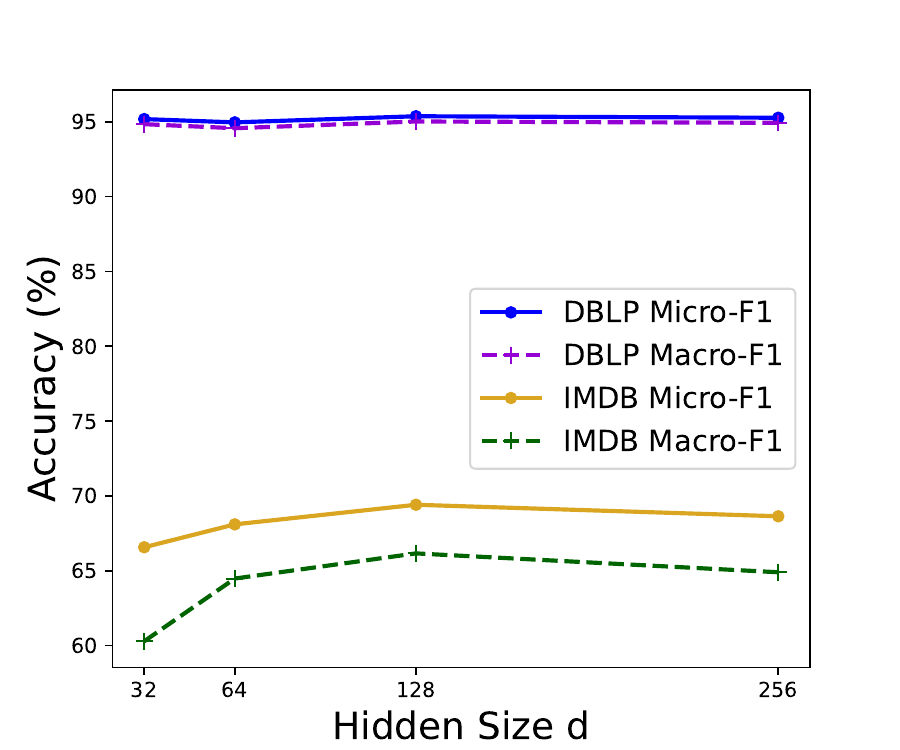}
    \end{subfigure}
    \hspace{-1mm}
    \caption{Hyper-parameters sensitivity studies.}
    \label{Hyperparameters}
\end{figure}
We also examine the sensitivity of hyperparameters, including the number of type-aware layers ($L$), dimension-aware layers ($L_d$), and hidden size. The results are depicted in Fig~\ref{Hyperparameters}.
We consistently observe strong performance across a wide range of $L_d$ values on both DBLP and IMDB datasets.
The impact of hidden size ($d$) is more significant on IMDB than on DBLP.
Increasing $L$ initially improves performance gradually, but further increments eventually lead to a decline, indicating potential harm from overly deep layers. 

\section{Conclusion}
In this paper, we investigate the problem of exploiting graph heterogeneity and the high-order feature information.
To achieve our goals, we propose {\name} composed of multiple cascade blocks, where each block comprises multiple type-aware layers and dimension-aware layers.
The type-aware encoder seamlessly integrates node types with node features to comprehensively leverage graph heterogeneity.  
The dimension-aware encoder pays attention to latent interactions among node features, utilizing the high-order information inherent in such interactions through a transformer architecture.
Extensive experiments and studies demonstrate the superiority, efficiency and robustness of the proposed {\name}.



%
%
%
\bibliographystyle{splncs04}
\bibliography{main}
%





\end{document}